\documentclass[runningheads]{llncs}
\usepackage{subcaption}
\usepackage{graphicx}
\usepackage{multirow}
\usepackage{amsmath,amssymb} 
\usepackage{color}
\usepackage[width=122mm,left=12mm,paperwidth=146mm,height=193mm,top=12mm,paperheight=217mm]{geometry}
\begin{document}
\pagestyle{headings}
\mainmatter

\title{Self-supervised Knowledge Distillation Using Singular Value Decomposition} 

\titlerunning{Self-supervised Knowledge Distillation Using Singular Value Decomposition}

\authorrunning{Seung Hyun Lee, Dae Ha Kim, Byung Cheol Song}

\author{Seung Hyun Lee\orcidID{0000-0001-7139-1764} 
Dae Ha Kim\orcidID{0000-0003-3838-126X} \and
Byung Cheol Song\orcidID{0000-0001-8742-3433}}


\institute{Inha	University, Incheon, Republic of Korea\\
	\email{ \{lsh910703, kdhht5022, \}@gmail.com, bcsong@inha.ac.kr}
}

\maketitle

\begin{abstract}
To solve deep neural network (DNN)'s huge training dataset and its high computation issue, so-called teacher-student (T-S) DNN which transfers the knowledge of T-DNN to S-DNN has been proposed. However, the existing T-S-DNN has limited range of use, and the knowledge of T-DNN is insufficiently transferred to S-DNN. To improve the quality of the transferred knowledge from T-DNN, we propose a new knowledge distillation using singular value decomposition (SVD). In addition, we define a knowledge transfer as a self-supervised task and suggest a way to continuously receive information from T-DNN. Simulation results show that a S-DNN with a computational cost of 1/5 of the T-DNN can be up to 1.1\% better than the T-DNN in terms of classification accuracy. Also assuming the same computational cost, our S-DNN outperforms the S-DNN driven by the state-of-the-art distillation with a performance advantage of 1.79\%. code is available on https://github.com/sseung0703/SSKD\_SVD. 
\keywords{Statistical methods and learning, Optimization methods, Recognition: detection, categorization, indexing, matching}
\end{abstract}

\section{Introduction}
Recently, DNN has overwhelmed other machine learning methods in the research fields such as classification and recognition
\cite{lecun1998gradient,krizhevsky2012imagenet}. As a result of the development of general-purpose graphics processing unit (GP-GPU) with high computational power, DNNs with huge complexity can be implemented and verified, resulting in DNNs that are superior to human recognition capabilities \cite{he2016deep,huang2017densely,xie2017aggregated}. On the other hand, it is still challenging to operate DNN on a mobile device or embedded system due to limited memory and computational capability. Recently, various lightweight DNN models have been proposed to reduce memory burden and computation cost \cite{zhang2017shufflenet,howard2017mobilenets}. However, these small-size models have less performance than state-of-the-art models like ResNext \cite{xie2017aggregated}. Another problem is that not only the conventional DNN but also the lightweight DNN model requires huge data in learning.

As a solution to these two problems, Hinton et al. \cite{hinton2015distilling} defined the concept of knowledge distillation and presented a teacher-student (T-S) DNN based on it. Then several knowledge distillation techniques have been studied \cite{romero2014fitnets,yim2017gift}. For example, in \cite{yim2017gift}, Yim et al. proposed a method to transfer the correlation between specific feature maps generated by T-DNN as the knowledge of T-DNN to the S-DNN. In this case, the S-DNN learns in two stages: the first stage that initializes the network parameters using the transferred knowledge, and the second stage that learns the main task.

However, the existing T-S knowledge distillation approaches have several limitations as follows: (1) They do not yet extract and distill rich information from the T-DNN. (2) In addition, the structure of T-S-DNN is very limited. (3) Finally, since the knowledge from the T-DNN is learned only for the purpose of initializing the parameters of the S-DNN, it gradually disappears as the learning of the next main task progresses.

In order to solve this problem, this paper approaches two perspectives. The first is a proper manipulation of knowledge for smaller memory and lower computation. So we gracefully compress the knowledge data by utilizing singular value decomposition (SVD), which is mainly applied to dimension reduction of features \cite{alter2000singular,zhang2014novel,ionescu2015matrix} in signal processing domain. We also analyze the correlation between compressed feature maps through a radial basis function (RBF) \cite{kim2006learning,wang2006using}, which is often used for kernelized learning. As a result, knowledge distillation using SVD and RBF can distill the information of T-DNN more efficiently than conventional techniques, and can transfer regardless of the spatial resolution of feature maps. Second, the training mechanism \cite{larsson2016learning,noroozi2016unsupervised,doersch2017multi} through self-supervised learning, which learns to create labels by itself, ensures that the transferred knowledge does not vanish and is continuously used. That is, it can figure out the vanishing problem of T-DNN knowledge. In addition, self-supervised learning can be expected to provide additional performance improvement because it allows for more powerful regularization \cite{hinton2015distilling}.

The experimental results show that when the visual geometry group (VGG) model \cite{simonyan2014very} is applied to the proposed network, T-DNN with 64.4\% accuracy for CIFAR-100 can improve the performance of S-DNN with 1/5 computation cost of T-DNN by 65.1\%. In addition to VGG, state-of-the-art models such as MobileNet\cite{howard2017mobilenets} and ResNext\cite{xie2017aggregated} are also applied to the proposed knowledge distillation method, confirming similar effects and proving that the proposed method can be generalized. Finally, we introduced self-supervised learning to continuously deliver the T-DNN's knowledge. As a result, we confirmed that the performance of the S-DNN is further improved by a maximum of 1.2\%, and finally the performance of the S-DNN becomes superior to the T-DNN by 1.79\%.

\section{Related Works}
    \subsection{Knowledge Distillation}
    Knowledge transfer is a technique for transferring information from a relatively complex and deep model, i.e., T-DNN to a smaller DNN model, i.e., S-DNN, ultimately increasing the performance of the S-DNN \cite{hinton2015distilling}. FitNet \cite{romero2014fitnets} first introduced the two-stage method to re-train the main task of the S-DNN after transferring knowledge of the T-DNN. The S-DNN could have much better initial parameters by learning knowledge distilled from the T-DNN than random initialization. Yim et al. \cite{yim2017gift} defined the knowledge transferred from the T-DNN to the S-DNN as changes of feature maps rather than layer parameters. They determined a certain layer group in the network and defined the correlation between input and output feature maps of the layer group as a Gram matrix so that the feature correlations of the S- and T-DNN become similar. However, the knowledge defined by the above techniques still lacks information, and knowledge transfer through initialization is still limited.

    \subsection{SVD and RBF}
    SVD is mainly used for dimension reduction or for extracting important information from feature maps \cite{alter2000singular,zhang2014novel,ionescu2015matrix}. In \cite{alter2000singular}, Alter et al. showed that it is possible to abstract the information of a dataset by using SVD. Lonescu et al. defined the gradient according to the chain rule for SVD, and proved that end-to-end learning is realizable even in DNN using SVD \cite{ionescu2015matrix}. They also showed that pooling high-level information in the feature map is very effective in the feature analysis tasks such as recognition and segmentation.
    RBF is a function that re-maps each feature in a viewpoint of distance from the center so that the feature has a high dimension. RBF can be used for various kernelized learning or RBF network (RBFN) \cite{kim2006learning,wang2006using}. In particular, analyzing features with RBF such as Gaussian function makes it possible to analyze noisy data more robustly.
    If these two methods can be combined well, it will be possible to extract important information effectively from fuzzy and noisy data. The proposed knowledge distillation method efficiently extracts core knowledge from a given feature map using SVD and effectively computes the correlation between two feature maps using RBF.

    \subsection{Training Mechanism}
    Self-supervised learning generates labels and learns them by itself. Recently, various self-supervised learning tasks have been studied \cite{larsson2016learning,noroozi2016unsupervised,doersch2017multi} because they can effectively initialize the network model. In \cite{doersch2017multi}, a method to learn various self-supervised tasks at a time by bundling them into a multi-task has been proposed and proved to be more efficient than conventional methods.
    On the other hand, semi-supervised learning is another learning scheme that uses labeled and unlabeled data at the same time when labeling data is insufficient. In order to solve the fundamental problem of the lack of a training-purpose dataset, various studies on semi-supervised learning have been actively conducted \cite{zhou2014semi,su2016efficient}.
    
    We will introduce the above-mentioned self-supervised learning as a more efficient transfer approach than parameter initialization through knowledge transfer in the existing T-S-DNNs.

\begin{figure}[t]
\centering
\includegraphics[width=11cm]{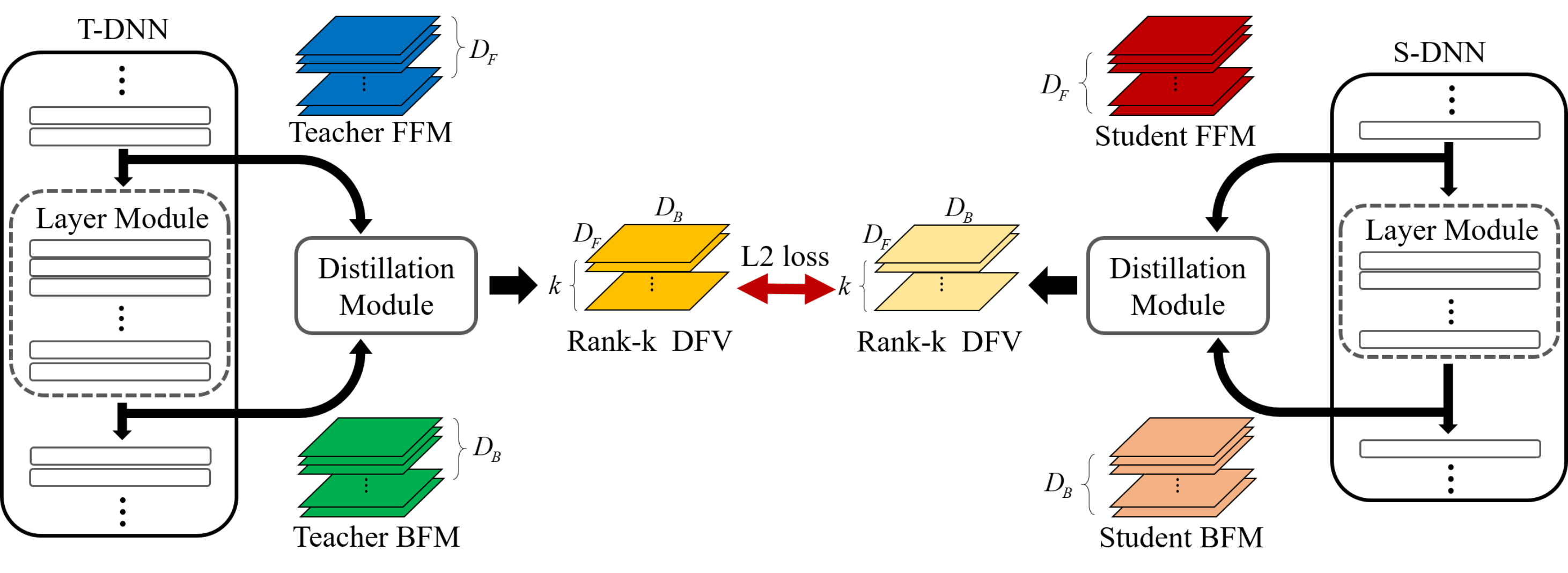}
\caption{The concept of the proposed knowledge distillation-based network.}
\label{fig:concept}
\end{figure}

\section{Method}
This section details the proposed knowledge transfer method. Inspired by the idea of \cite{yim2017gift}, we derive a correlation between two feature maps extracted from T-DNN, and transfer it as knowledge. Fig.~\ref{fig:concept} illustrates the proposed knowledge distillation based network. First, both the T-DNN and the S-DNN are composed of a predetermined convolutional layer and a fully-connected layer depending on the purpose. For example, VGG \cite{simonyan2014very}, MobileNet \cite{howard2017mobilenets}, ResNext \cite{xie2017aggregated}, etc. can be adopted as DNN. Then, to extract the feature map characteristic inherent to each DNN, we specify two particular layer points in the DNN and sense the corresponding two feature maps. The layers between the two points are defined as a layer module. The feature map that is sensed at the input of the layer module is called the front-end feature map (FFM) and the feature map that is sensed at the output is called the back-end feature map (BFM). For example, in MobileNet, the layer module can consist of several depth-wise separable convolutions. Let the depths of FFM and BFM be $D_{F}$ and $D_{B}$, respectively. On the other hand, several non-overlapping layer modules may be defined in each DNN for robust distillation. In this paper, the maximum number of layer modules in each DNN is $G$.

Now we can get the correlation between FFM and BFM of a certain layer module through the distillation module. The distillation module outputs the distillation feature vectors (DFV) having the size of $k \times D_{F} \times D_{B}$ from two inputs of FFM and BFM. See Sec.~\ref{sec:module}.

Finally, we propose a novel training mechanism so that the knowledge from the T-DNN does not disappear in the 2nd stage, i.e., main-task learning process. We improve self-supervised learning, which was presented in \cite{hinton2015distilling}, to enable more effective transfer of knowledge. See Sec. 3.2.

\subsection{Proposed Distillation Module}\label{sec:module}
    In general, DNNs generate feature maps through multiple layers to suit a given task. In the distillation method of \cite{yim2017gift}, the correlation between feature maps obtained from DNN is first defined as knowledge. The proposed method also accepts the idea of \cite{yim2017gift} and distillates the knowledge using correlation between feature maps. However, feature maps that are produced through multiple convolution layers are generally too large to be used as they are not only computationally expensive, but also difficult to learn. An intuitive way to solve this problem is to reduce the spatial dimensions of the feature maps. We introduce SVD to effectively remove spatial redundancy in feature maps and obtain meaningfully implied feature information in the process of reducing feature dimensions. This section describes in detail how to generate DFV, i.e., knowledge for distillation using SVD.
    
    Fig. 2 shows the structure of the proposed knowledge distillation module. Suppose that the input and output feature maps of the layer module defined in T-DNN, i.e., FFM and BFM are inputs to this distillation module. First, we eliminate the spatial redundancy of feature maps by using truncated SVD. Then, the right-hand singular vectors V obtained from the truncated SVD and the singular value matrix   are post-processed for easy learning, and then $k$ feature vectors are obtained. Finally, the correlation between feature vectors obtained from FFM and BFM is computed by RBF to obtain a rank-$k$ DFV.
    
    \begin{figure}[t]\centering
    \includegraphics[width=8.0cm]{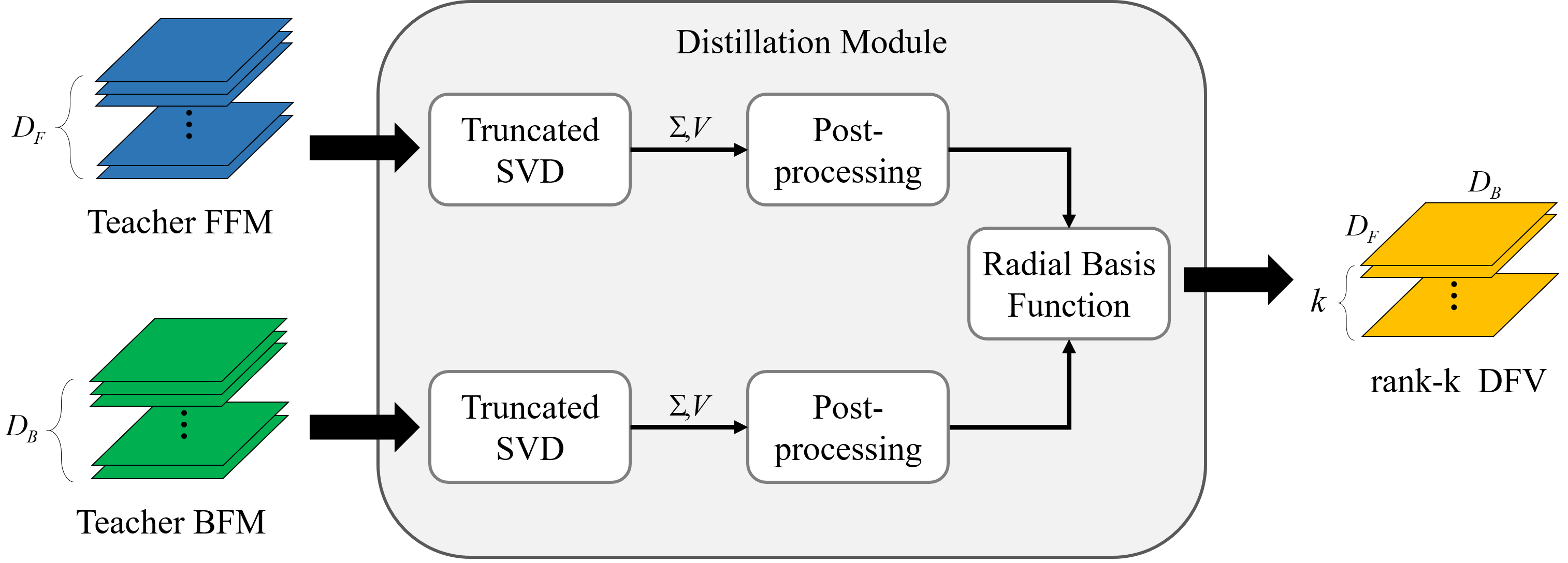}
    \caption{The proposed knowledge distillation module.}
    \label{fig:module}\end{figure}
    
\subsubsection{Truncated SVD} 
    As shown in Fig. 3(a), the first step of the distillation module is the truncated SVD which is used to compress the feature map information and lower the dimension simultaneously. Prior to applying SVD, preprocessing is performed to convert the 3D feature map information of $H\times W\times D$ into a 2D matrix $M$ having $\left(H \times W\right)\times D$ size. Then $M$ can be a factorization of the form $U\Sigma V^{T}$ by SVD. $V^{T}$ is the conjugate transpose of $V$. The columns of $U$ and the columns of $V$ are called the left-singular vectors and right-singular vectors of $M$, respectively. The non-zero singular values of $M$ (found on the diagonal entries of $\Sigma$) are the square roots of the non-zero eigenvalues of both $M^{T}M$ and $MM^{T}$. On the other hand, $U$ and $V$ decomposed through SVD have different information \cite{alter2000singular}. $U$ is the unique pattern information of each feature of $M$, and $V$ can be interpreted as global information of the feature set. And $\Sigma$ has the scale or energy information of the singular value. Since we aim to obtain compressed feature information, we use only $V$ having global information of the feature map and its energy $\Sigma$.
    
    To minimize memory size as well as computational cost, we use truncated SVD. Truncated SVD refers to an SVD that decomposes a given matrix by only a pre-determined rank $k$. That is, $V$ and $\Sigma$ have dimensions of $k\times D$ and $k \times 1$, respectively. In this case, since the difference between the re-composed matrix and the original matrix is minimized, the information of the given matrix $M$ can be maintained as much as possible. As a result, FFM and BFM are compressed with minimal loss of information as shown on Fig. 3(a).
    
    On the other hand, in order to apply the chain rule by back propagation to the truncated SVD part in the learning process, the gradient of $M$ must be defined. So, we modify the gradient defined in \cite{ionescu2015matrix}. Note that the proposed scheme uses only $V$ and $\Sigma$ among decomposed vectors, unlike \cite{ionescu2015matrix}. Since $\Sigma$ is simply used as a scale factor, it is not necessary to obtain its gradient. Therefore, only the gradient for $V$ is obtained and the gradient of $M$ is re-defined as in Eq. (1) to Eq. (2).
    
    \begin{align}\label{eq:svd_grad0}
    \nabla\left(M \right) = \left\{\begin{matrix}
    UE^{T}&-U\left(E^{T}V \right)_{diag}V^{T}&                                                              &\multirow{2}{*}{$HW \leq D$}\\
          &\multicolumn{2}{c}{-2U\left(K\circ\left(\Sigma ^{T} V^{T}E \right )\right)_{sym}\Sigma^{T}V^{T},}& \\
    \\
    \multicolumn{3}{c}{2U\Sigma \left(K^{T} \circ\left(V^{T} \nabla \left(V \right ) \right ) \right )_{sym}V^{T},} & otherwise\\
    \end{matrix}\right.
    \end{align}

    \begin{align}\label{eq:svd_grad}
    E = \nabla\left( V \right)\Sigma^{-1}
    , K = \left\{\begin{matrix}
    \frac{1}{\sigma_{i}^{2}-\sigma_{j}^{2}}, & i \neq j,&\left(1\leq i,j \leq k\right) \\ 
                                           0,& \multicolumn{2}{c}{otherwise}
    \end{matrix}\right.
    \end{align}
    where $\left(A \right)_{sym}=\frac{1}{2}\left(A^{T}+A \right)$and $\left(A \right)_{diag}$ is a function that makes all off-diagonal components zero. Also $\circ$ indicates Hadamard product, and $\sigma$ stands for diagonal component of $\Sigma$ . We do not need to perform unnecessary operations on $\nabla\left(\Sigma \right)$ and $\nabla\left(U \right)$, and since the dimension of each matrix is low, the computation cost can be minimized as a whole.
    
    Therefore, truncated SVD is a key element of the proposed knowledge distillation module because it effectively reduces the dimension of the feature map. As a result, the proposed knowledge distillation functions to fit the small size network.

    \begin{figure}[t]
    \centering
    \begin{subfigure}[b]{0.3\textwidth}
    \centering
    \includegraphics[height=4cm]{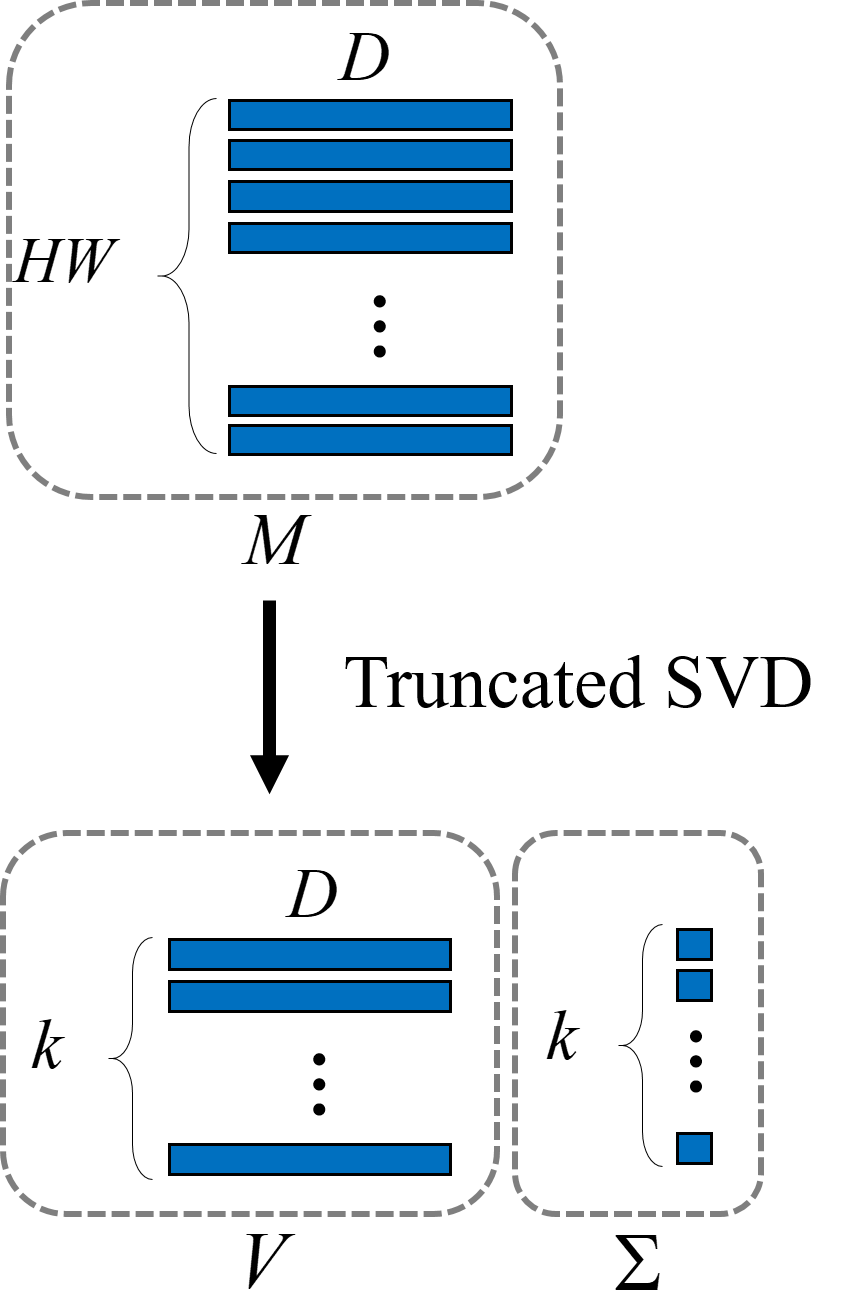}
    \caption{(a)}
    \end{subfigure}
    \begin{subfigure}[b]{0.3\textwidth}
    \centering
    \includegraphics[height=4cm]{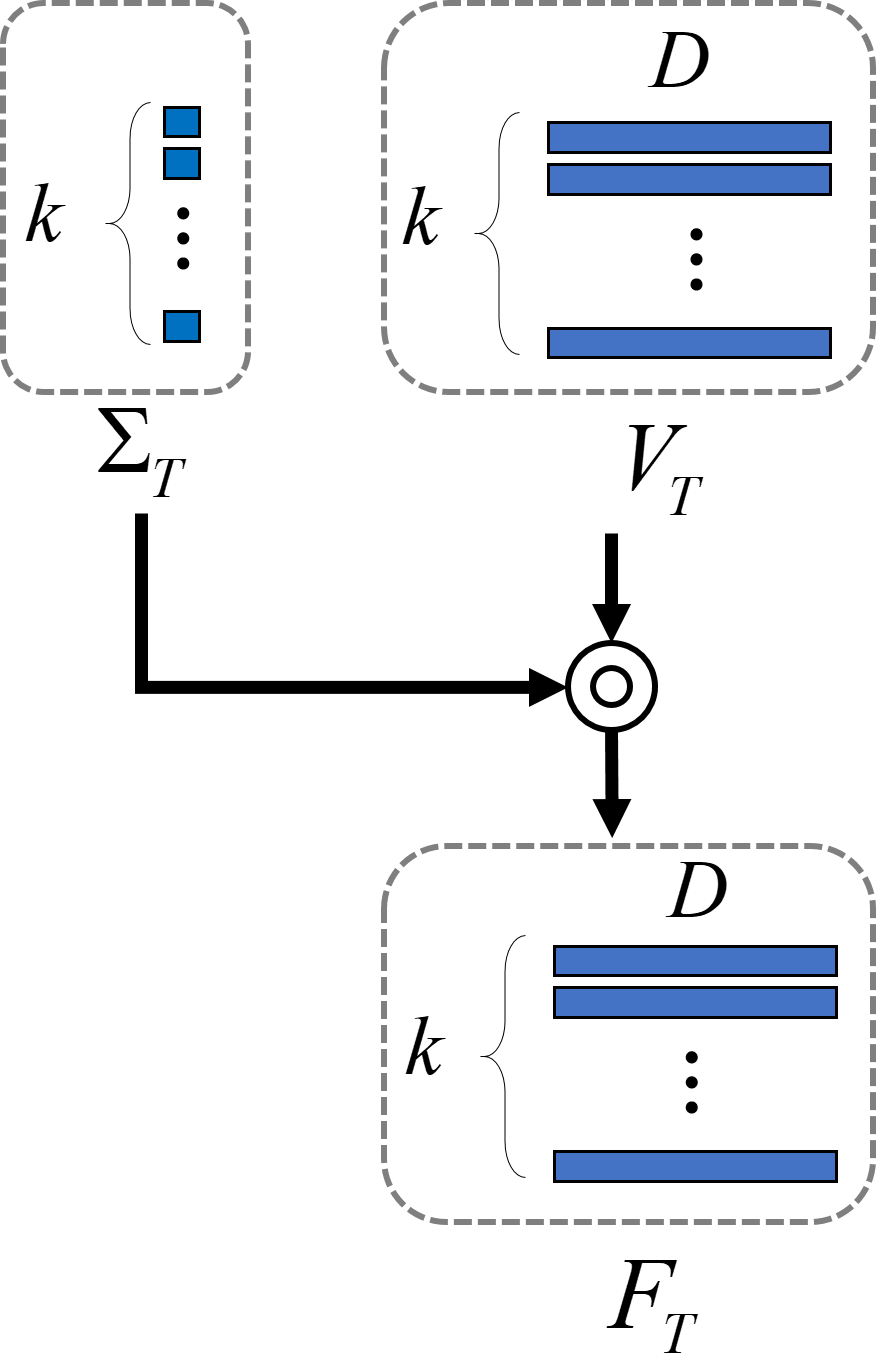}
    \caption{(b)}
    \end{subfigure}
    \begin{subfigure}[b]{0.3\textwidth}
    \centering
    \includegraphics[height=4cm]{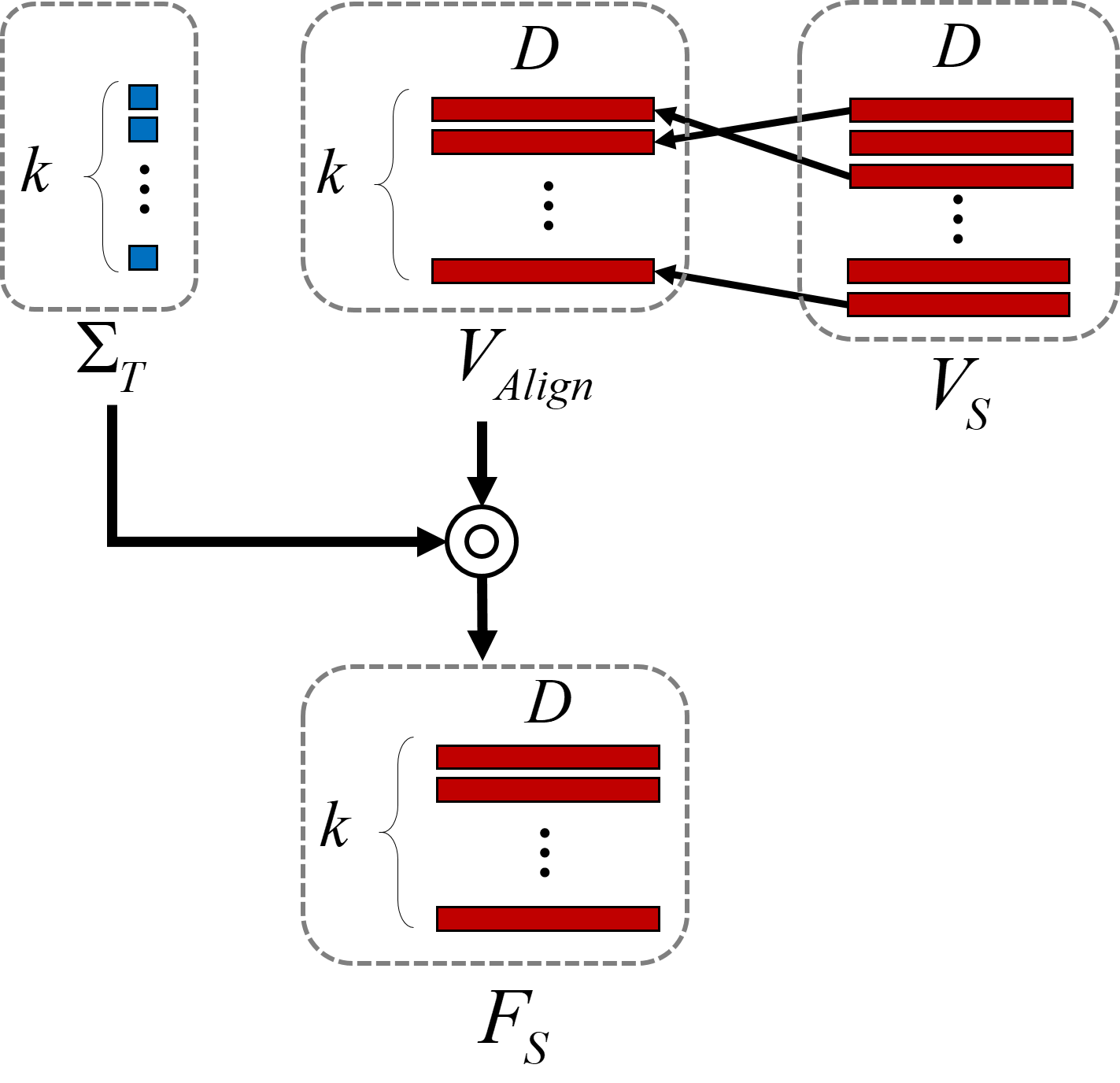}
    \caption{(c)}
    \end{subfigure}
    \setcounter{figure}{2}
    \caption {(a) Truncated SVD (b) post-processing of T-DNN (c) post-processing of S-DNN}
    \end{figure}

\subsubsection{Post-processing}
    Truncated SVD products, $V$ and $\Sigma$ contain enough FFM and BFM information, but are difficult to use directly because of the following two problems. First, since SVD decomposes a given matrix in decreasing order of energy, the order of singular vectors with similar energy can be reversed. Second, because each element of the singular vector has a value of [-1,1], singular vectors with the same information but the opposite direction may exist. So, even with similar feature maps, the results of decomposing them may seem to be very different.
    
    Therefore, the corresponding singular vectors of T-DNN and S-DNN are post-processed differently based on T-DNN because T-DNN delivers its information to S-DNN. First, post-processing for T-DNN is described in Fig. 3(b). The singular value of T-DNN $\Sigma_{T}$ is normalized so that the square sum becomes 1. Normalization is performed by multiplying a normalized $\Sigma_{T}$ with singular vector of T-DNN $V_{T}$ as shown in Eq. (4) to obtain a set of compressed feature vectors $F_{T}$ as shown in Eq. (3).
    
    \begin{align}
    & F_{T}=\left\{ f_{T,i} | 1 \leq i \leq k \right\}
    \end{align}
    \begin{align}
    f_{T,i}=\frac{\sigma_{T,i} }{\left|| \Sigma_{T} \right||_{2}}v_{T,i}
    \end{align}
    where $\sigma_{T,i}$ is the $i$-th singular value of T-DNN and $v_{T,i}$ is the corresponding singular vector. Since the singular value means the energy of the corresponding singular vector, each singular vector is learned in order of importance.
    
    Next, a singular vector of S-DNN is post-processed as shown in Fig. 3(c). First, we align the student singular vectors based on the teacher singular values. So the student singular vector with the most similar information to the teacher singular vector is aligned in the same order.
    
    Here, the similarity between singular vectors is defined as the absolute value of cosine similarity, which determines the similarity degree through the angles between two vectors so that the similarity between the vectors with opposite directions can be accurately measured. This process is described in Eqs. (5-6).
    \begin{align}
    s_{j}=\underset{j}{\textrm{argmax}}\left( \left| v_{T,i}\cdot v_{S,j}\right |\right), & \left(1\leq i\leq k \right),\left(1 \leq j \leq k+1 \right)
    \end{align}
    \begin{align}
    v_{Align,i} = v_{S,s_{j}}
    \end{align}
    Here $v_{S,j}$ indicates the $j$-th vector of the S-DNN's $V$ and $v_{Align,i}$ is the $i$-th vector of the aligned version of the S-DNN's $V$. Note that for effective alignment, the student feature map decomposes one more vector. Also, the singular vectors of S-DNN are normalized by the singular values of T-DNN, so that a singular vector of higher importance is further learned. This is shown in Eqs. (7-8).
    \begin{align}
    F_{S}=\left\{ f_{S,i} | 1 \leq i \leq k \right\}, 
    \end{align}
    \begin{align}
    f_{S,i}=\frac{\sigma_{T,i} }{\left|| \Sigma_{T} \right||_{2}}v_{Align,i}
    \end{align}
    Thus, because of the post-processing, noisy and randomly decomposed singular vector information can be used effectively.\\
    
\subsubsection{Computing Correlation using Radial Basis Function}
    This section describes the process of defining knowledge by the correlation of the feature vectors obtained in the previous section. Since the derived feature information from a singular vector is generally noisy, noise-robust methods are required. Therefore, we employ Gaussian RBF, which is a frequently used kernel function for analyzing noisy data \cite{kim2006learning,wang2006using}, as a way to obtain the correlation.
    
    On the other hand, feature vectors obtained by applying the proposed SVD and post-processing to FFM and BFM are basically discrete random vectors independent of each other. Thus, we define the correlation between feature vector sets obtained from FFM and BFM as a point-wise $L_{2}$ distance as in Eq. (10), and the rank-$k$ DFV are completed by applying Gaussian RBF to the computed correlation as in Eq. (9) for the dimension extension.
    
    \begin{align}
    DFV =  \left \{\textrm{exp}\left( -\frac{d_{m,n,l}}{\beta}\right),1\leq m \leq D_{F},1\leq n \leq D_{B},1\leq l \leq k \right\}
    \end{align}
    
    \begin{align}
    d_{m,n,l} = \left\|f_{m,l}^{FFM}-f_{n,l}^{BFM}\right\|_{2}^{2}
    \end{align}
    
    $\beta$ in Eq. (9) is a hyper-parameter for smoothing DFV and it should be properly selected for noise-robust operation.
      
    As mentioned above, the correlation between feature maps composed of noisy and fuzzy data can be effectively obtained through SVD and RBF. Therefore, the distillated knowledge from T-DNN by the proposed scheme can be a very effective guidance for S-DNN. Also, unlike the existing technique, DFV can transfer knowledge regardless of feature map size and therefore it causes consistent performance. The experimental results are discussed in Section 4.2.

\subsection{Training Mechanism}
    The remaining step is to learn to improve the performance of S-DNN by transferring distilled knowledge of T-DNN, i.e., DFV, to S-DNN. We need to learn that the S-DNN imitates the T-DNN with the DFV as an intermediary, so we define the $L_{2}$ loss function $L_{transfer}\left( DFV_{T},DFV_{S}\right)$ of the knowledge pair of T-DNN and S-DNN as Eq. (11).
    \begin{align}
    L_{transfer}\left( DFV_{T},DFV_{S}\right) = \sum_{g}^{G}\frac{\left\| DFV_{T}^{\left(g\right)}-DFV_{S}^{\left(g\right)} \right\|_{2}^{2}}{2}
    \end{align}
    where $G$ is the maximum number of layer modules defined in the proposed T-S-DNN. In this case, all layer modules are assumed to have the equivalent importance, and are trained without additional weighting. If S-DNN is initialized by transferring knowledge of T-DNN to S-DNN through learning based on Eq. (11), the learning performance of the main task of S-DNN can be improved (see Section 4.2).
    
    However, even though learning the main task of S-DNN after initialization as described above, there is still a problem that the knowledge of T-DNN gradually disappears as learning progresses and the performance improvement is limited. So we introduce self-supervised learning to train both main task and transfer task at the same time. Since the knowledge of T-DNNs learned by S-DNN is a label generated by T-DNN, self-supervised learning is possible using this characteristic. As a result, the final loss function for learning the parameter of S-DNN $\Theta_{S}$ is defined as Eq. (12).
    \begin{align}
    L_{total}\left( \Theta_{S} \right)=L_{main}\left( \Theta_{S} \right)+L_{transfer}\left( DFV_{T},DFV_{S} \right)
    \end{align}
    As described above, when the main task and the transfer task are learned together by a multi-task learning, it is possible to continuously transfer knowledge of T-DNN to further improve the performance.
    
    On the other hand, if the distillation loss is much larger than the main task loss, the gradient of knowledge transfer becomes too large and the above multi-task learning may not work properly. To solve this problem, it is necessary to limit the effect of the distillation task. So we introduce a gradient clipping \cite{pascanu2013difficulty} to limit the gradient of knowledge transfer.
    
    In general, the threshold for clipping is constant, but we define the $L_{2}$-norm ratios of the main task and the transfer task as shown in Eq. (13), and clip the gradient of the knowledge transfer adaptively using this. In addition, since randomly initialized S-DNN is different from T-DNN, it is difficult to follow T-DNN fast. Therefore, we use a sigmoid function as shown in Eq. (14) to design the clipped gradient to grow smoothly as learning progresses.
    
    \begin{align}
    \tau = \frac{\left\|\nabla\left(\Theta_{S} \right)_{main} \right\|_{2}}{\left\|\nabla\left(\Theta_{S} \right)_{trans} \right\|_{2}}
    \end{align}
    \begin{align}
    \nabla\left(\Theta_{S} \right )_{trans}^{clipped} = \left\{\begin{matrix}
    \frac{1}{1+\textrm{exp}\left(-\tau + p \right)}\nabla\left(\Theta_{S} \right)_{trans}, & \nabla\left(\Theta_{S} \right )_{trans} < \nabla\left(\Theta_{S} \right )_{main}\\
    \nabla\left(\Theta_{S} \right )_{trans} & otherwise
    \end{matrix}\right.
    \end{align}
    
    In Eq. (14), $p$ means the current epoch. Therefore, the proposed self-supervised learning method can concentrate more on the learning of the main task while learning the two tasks of different nature at the same time. In other words, rich knowledge distillated from T-DNN can be continuously transferred to S-DNN without vanishing. In addition, since the proposed self-supervised learning method has the effect of hard regularization of S-DNN, the performance of S-DNN can be improved without over-fitting (see Section 4.3).

\section{Experimental Results}
    In order to evaluate the performance of the proposed knowledge distillation method, we performed the following three experiments. First, we verified the effectiveness of the proposed knowledge itself. To do this, we conducted experiments on so-called small network enhancement that improves the performance of a relatively small S-DNN using T-DNN knowledge (see Section 4.2). Second, we examined the performance of the training mechanism proposed in Section 3.2 (see Section 4.3). Here, the comparison target was Yim et al.'s two-stage approach \cite{yim2017gift}. Section 4.3 also demonstrates that the proposed method can transfer knowledge robustly even when there is no hard constraint on image information. Third, the performance of the proposed method according to the number of DFVs is experimentally examined in Section 4.4.

\subsection{Experiment Environments}
    We implemented the proposed method using Tensorflow \cite{tensorflow2015-whitepaper} on a computer with specification of the Intel짰 Core i7-7700 CPU@3.60GHz x8, 16GB RAM, and GeForce GTX 1070. We used two datasets, CIFAR100 \cite{krizhevsky2009learning} and Imagenet32 \cite{chrabaszcz2017downsampled}. The CIFAR100 dataset consists of color images with a small size of 32x32, with 50,000 training data and 10,000 test data divided into 100 categories or labels. Imagenet32 is a huge dataset made up of small images called the down-sampled version of Imagenet \cite{imagenet_cvpr09}. Imagenet32 is composed of 1,281,167 training data and 50,000 test data with 1,000 labels. We augmented the dataset to verify the regularization performance of the proposed method. The augmentations used here are random shift, random rotation, and horizontal flip. The proposed method was tested under the same conditions as \cite{yim2017gift}, and the average of three equivalent experimental results was used as the final result to increase the reliability of the results.

\subsection{Small Network Enhancement}
    \begin{figure}[t]
    \centering
    \begin{subfigure}[b]{0.3\textwidth}
    \centering
    \includegraphics[height=4.5cm]{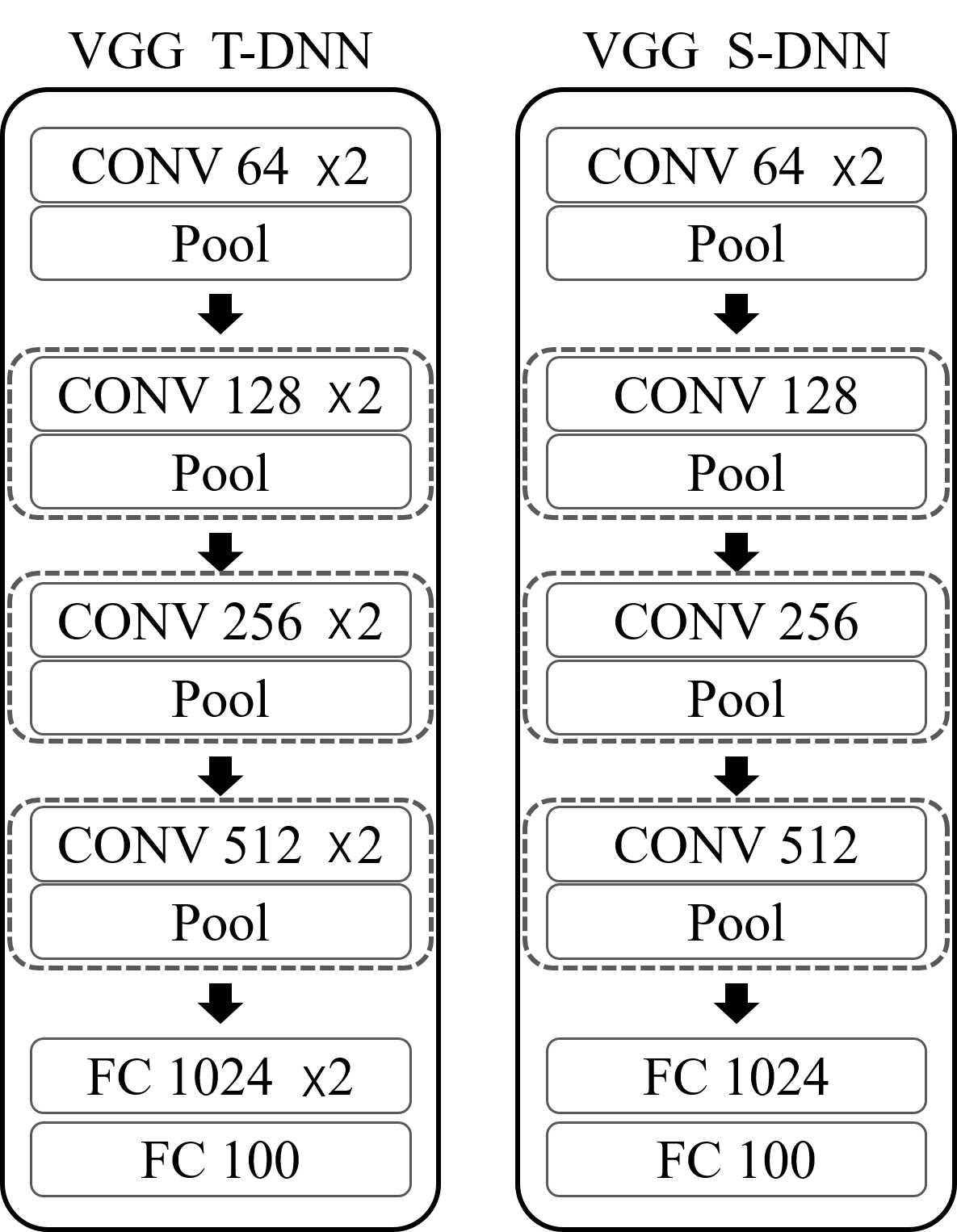}
    \caption{(a)}
    \end{subfigure}
    \begin{subfigure}[b]{0.3\textwidth}
    \centering
    \includegraphics[height=4.5cm]{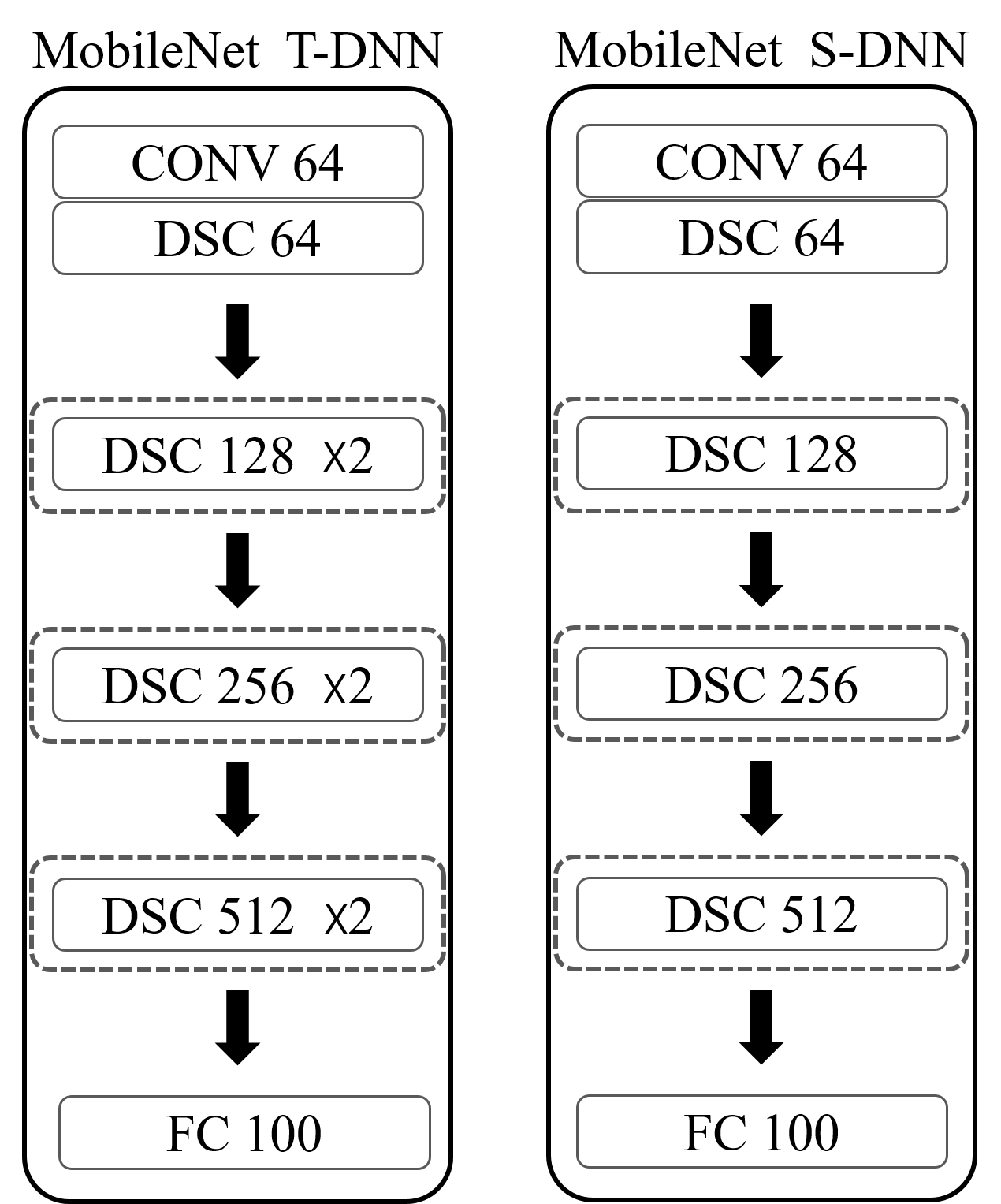}
    \caption{(b)}
    \end{subfigure}
    \begin{subfigure}[b]{0.3\textwidth}
    \centering
    \includegraphics[height=4.5cm]{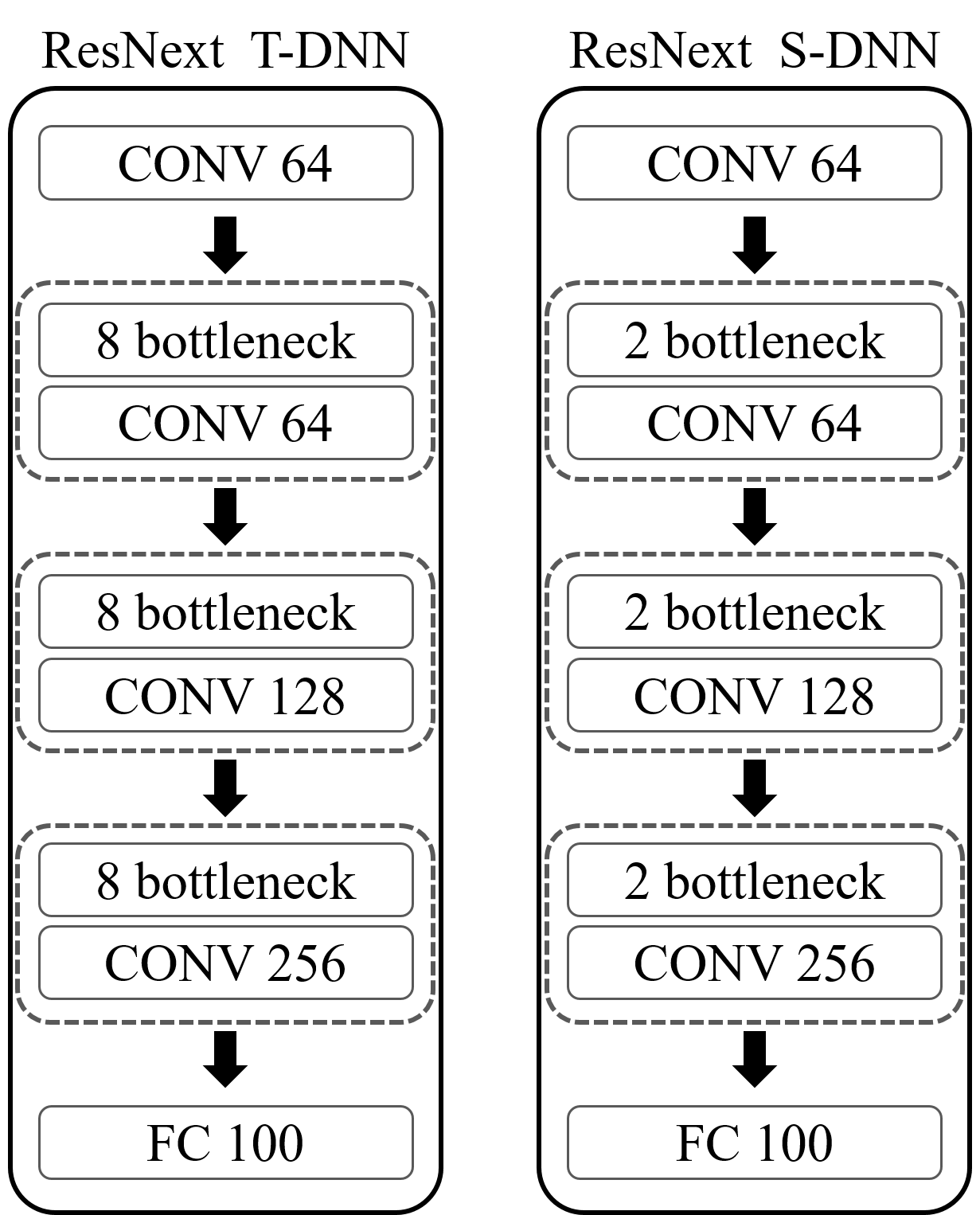}
    \caption{(c)}
    \end{subfigure}
    \setcounter{figure}{3}
    \caption {A pair of T-DNN and S-DNN for an experiment to evaluate small network enhancement. (a) VGG, (b) MobileNet, (c) ResNext. Here dotted boxes indicate layer modules.}
    \end{figure}
    
    In order to verify the effect of knowledge transfer only, we first showed the result of learning in two-stage approach as in \cite{yim2017gift}. That is, the self-supervised learning of Section 3.2 was not used in this experiment. We compared the proposed method and the state-of-the-art knowledge distillation method \cite{yim2017gift}. In addition, the results of T-DNN alone and S-DNN alone were also shown. All the methods were learned with the CIFAR100 dataset. We employed VGG, MobileNet, ResNext as the DNN to apply to the proposed method. The T-S-DNNs constructed using these are shown in Fig. 4.
    
    Although VGG is somewhat poorer than the state-of-the-art CNN models in terms of ratio of accuracy and parameter size, it is widely used because of its simple structure and ease of implementation. We used a modified version of the T-DNN for CIFAR100 by removing the last three convolutional layers from the VGG network proposed in \cite{simonyan2014very}. The S-DNN consists of only one convolutional layer with the same filter depth as shown in Fig. 4(a). Here, the layer module is defined as a convolutional layer with the same filter depth.
    
    MobileNet is a CNN with small parameter size and computational cost designed for use in mobile or embedded environments. The MobileNet case shows that the proposed method is capable of improving performance even for small networks. As shown in Fig. 4(b), T-DNN was constructed by removing the last four depth-wise separable convolutional layers (DSC) proposed in \cite{howard2017mobilenets} to fit CIFAR100. The S-DNN is composed by using the DSC of the same filter depth only once. Here, the layer module is defined by the DSC of the same filter depth.
    
    Finally, ResNext is a network where the convolution layer was divided into several bottleneck layers. Through experiments using ResNext, we show that the proposed method can transfer knowledge effectively even in networks with very complex structures. We used the network proposed in \cite{xie2017aggregated} as the T-DNN and the S-DNN is constructed by partially reducing the bottleneck layers. Here, the layer module is defined by combining the bottleneck layer and one convolutional layer (see Fig. 4(c)).\\
    The weight of each network was determined by He's initialization \cite{he2016deep} and $L_{2}$ regularization. Decay parameter was set to $10^{-4}$. Batch size was set to 128, and stochastic gradient descent (SGD) \cite{kiefer1952stochastic} was used for optimization, and Nesterov accelerated gradient \cite{nesterov1983method} was applied. The initial learning rate was set to $10^{-2}$ and the momentum was set to 0.9. During a total of 200 epochs, the networks were learned and the learning rate was reduced to 1/10 per 50 epochs. Both stages used the same hyper-parameters. The hyper-parameter of the proposed method $k$ was set to is 1. In other words, only one DFV is used and $\beta$ of RBF is experimentally fixed to 8.
    
    The experimental results are shown in Table 1, and it can be seen that the proposed method is always better than \cite{yim2017gift}. In the case of VGG, the proposed method has an outstanding performance improvement of 3.68\% compared to S-DNN. It also shows about 0.49\% better performance than \cite{yim2017gift} and 0.61\% higher performance than T-DNN alone. In case of Mobilenet, the proposed method improves the performance by about 2\% over S-DNN, and 1.62\% over \cite{yim2017gift} and 0.3\% over T-DNN. This shows that the proposed method is more suitable for small networks than \cite{yim2017gift}. In the case of ResNext, the proposed method improves the performance of S-DNN by only 1.43\%, which is lower than that of VGG or MobileNet, but has a performance advantage over 1.83\% than \cite{yim2017gift}. This result shows that the proposed method works well in a state-of-the-art network with a complicated structure such as ResNext. Therefore, the proposed method effectively compresses knowledge of T-DNN and transfers the compressed knowledge regardless of network structure.
    \begin{table}[t]
    \begin{center}
    \caption{Comparison of the proposed algorithm with \cite{yim2017gift} for three different networks. Here, FLOPS indicates the sum of the numbers of addition, multiplication, and condition. ?쁏arams??indicates the sum of weights and biases.}
    \label{table:headings}
    \begin{tabular}{c|c|c|c|c}
    \hline\hline\noalign{\smallskip}
    Network & Model & FLOPs & Params & Accuracy\\
    \noalign{\smallskip}
    \hline\hline
    \noalign{\smallskip}
    \multirow{4}{*}{VGG}       & T-DNN             & 576.3M & 10.9M & 64.44 \\
                               & S-DNN             & 121.3M & 3.8M  & 61.37 \\
                               & \cite{yim2017gift}& 121.3M & 3.8M  &64.54 \\
                               & proposed          & 121.3M & 3.8M  & \textbf{65.05} \\
    \hline
    \multirow{4}{*}{MobileNet} & T-DNN             &  98.4M & 2.3M  & 57.85 \\
                               & S-DNN             &  37.8M & 0.82M & 56.15 \\
                               & \cite{yim2017gift}&  37.8M & 0.82M & 56.53 \\
                               & proposed          &  37.8M & 0.82M & \textbf{58.15} \\
    \hline
    \multirow{4}{*}{ResNext}   & T-DNN             & 547.3M & 0.66M & 66.58 \\
                               & S-DNN             & 247.6M & 0.34M & 64.00 \\
                               & \cite{yim2017gift}& 247.6M & 0.34M & 63.60 \\
                               & proposed          & 247.6M & 0.34M & \textbf{65.43}\\
    
    \hline
    \end{tabular}
    \end{center}
    \end{table}
    
    \begin{table}[t]
    \begin{center}
    \caption{Sensitivity of the proposed network to spatial resolution of feature map.}
    \label{table:headings}
    \begin{tabular}{c|c|c|c|c}
    \hline\hline\noalign{\smallskip}
    Network & Model & FLOPs & Params & Accuracy\\
    \noalign{\smallskip}
    \hline\hline
    \noalign{\smallskip}
    \multirow{3}{*}{VGG} & T-DNN           & 576.3M & 10.9M & 64.44 \\
                         & S-DNN           &  15.6M &  3.8M & 54.17 \\
                         & Proposed        &  15.6M &  3.8M & 61.15 \\
    
    \hline
    \end{tabular}
    \end{center}
    \end{table}
    
    On the other hand, we constructed another VGG-based S-DNN to show that the proposed method can transfer knowledge regardless of the resolution of feature maps. In the convolutional layer of the S-DNN used above, the padding was not performed and the size of the feature map was reduced by setting the stride of the convolutional layer to 2 instead of pooling. This dramatically reduces the spatial resolution of the feature map as it passes through the convolution layer. The hyper-parameters used for learning were the same as before.
    
    Since knowledge transfer using \cite{yim2017gift} is impossible in this T-S-DNN structure, Table 2 shows only the results of the proposed method. We can see that the performance of S-DNN with FLOPS of about 0.03 times that of T-DNN is improved by about 6.98\%. Therefore, the proposed method can effectively transfer the knowledge of T-DNN regardless of the spatial resolution of the feature map, and is effective for practical applications requiring small size DNNs.

\subsection{Training Mechanism}
    In this section, we evaluate the training mechanism proposed in Section 3.2. The network used for learning is the VGG-based T-S-DNN used in Section 4.2. The hyper-parameters are the same as those used in Section 4.2.
    
    Table 3 shows the experimental results. The performance improvement was 0.35\% when the proposed training mechanism was applied to \cite{yim2017gift}, and the performance improved by 0.49\% when the proposed training mechanism was applied together with the proposed knowledge distillation technique. This is because S-DNN is regularized continuously without vanishing of knowledge of T-DNN. In addition, since the number of epochs required for learning is reduced by half compared with the conventional two stage structure, the learning time can be shortened significantly. Therefore, using both the knowledge distillation technique and the training mechanism, the performance improvement is expected to be about 4.17\% higher than that of the S-DNN alone. In addition, the proposed method can improve performance up to 1\% than \cite{yim2017gift} and 1.1\% over T-DNN. Since the computation cost of S-DNN amounts to only 1/5 of that of T-DNN, we can see that S-DNN is well regularized by the proposed method.
    
    \begin{table}[t]
    \begin{center}
    \caption{Performance evaluation according to training mechanism.}
    \label{table:headings}
    \begin{tabular}{c|c|c}
    \hline\hline\noalign{\smallskip}
    Model & Mechanism & Accuracy\\
    \noalign{\smallskip}
    \hline\hline
    \noalign{\smallskip}
    \multirow{2}{*}{\cite{yim2017gift}} & 2 Stage & 64.54 \\
                                        & 1 stage & 64.89 \\
    \hline
    \multirow{2}{*}{Proposed}           & 2 Stage & 65.05 \\
                                        & 1 stage & \textbf{65.54}\\
    \hline
    \end{tabular}
    \end{center}
    \end{table}

\subsection{Semi-supervised Learning using Knowledge Distillation}
    Semi-supervised learning is a fusion of supervised learning and unsupervised learning, and is very effective when learning DNNs using a dataset lacking labeled data \cite{krizhevsky2009learning,chrabaszcz2017downsampled}. Since the proposed method is basically composed of self-supervised learning, we could adapt the proposed method to semi-supervised learning without difficulty. 
    In order to verify the effectiveness of semi-supervised learning based on the proposed method, we adopted the VGG-based T-DNN used in Section 4.2 and the S-DNN used the same network as T-DNN. First, T-DNN was trained with Imagenet32 to have general information. The hyper-parameters used in this learning were the same as those used in Section 4.2 and the batch size was adjusted to 512. Based on the trained knowledge of T-DNN, S-DNN is trained by the following semi-supervised learning.
    The dataset for semi-supervised learning was CIFAR100, and only a certain percentage of labeled data was used per label. The hyper-parameters for this experiment were the same as those used in Section 4.2, and the total number of epochs was doubled so that the learning process is fully converged. In order to relatively evaluate the performance of the proposed method, the result of learning the knowledge of \cite{yim2017gift} using the proposed training mechanism was also shown (see Table 4).
    Table 4 and Fig. 5 show the experimental results for semi-supervised learning. The performance of the proposed method was always superior to \cite{yim2017gift}. As the number of labeled data decreases, \cite{yim2017gift} abruptly degraded, but the performance degradation of the proposed method was not relatively severe. In other words, the proposed method can effectively learn dataset with less image constraint information than \cite{yim2017gift}.

    \begin{table}[t]
    \begin{center}
    \caption{The effect of semi-supervised learning according to the portion of labeled data. Here perQ indicates the usage of Q\%, and Full stands for 100\%.}
    \label{table:headings}
    \begin{tabular}{c|c|cccc}
    \hline\hline\noalign{\smallskip}
    Network & Model & per10 & per20 & per50 & Full\\
    \noalign{\smallskip}
    \hline\hline
    \noalign{\smallskip}
    \multirow{2}{*}{VGG} & \cite{yim2017gift} & 33.92 & 41.86 & 55.21 & 64.44 \\
                         & Proposed           & 49.68 & 55.43 & 60.01 & 64.44 \\
    \hline
    \end{tabular}
    \end{center}
    \end{table}
    
    \begin{figure}[t]\centering
    \includegraphics[height=4.0cm]{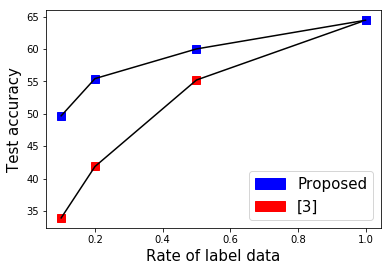}
    \caption{Test accuracy according to the portion of labeled data.}
    \label{fig:module}\end{figure}

\subsection{Performance Evaluation According to the Number of DFVs}
    The number of DFVs to be transferred in the proposed knowledge distillation has a significant impact on overall performance. For example, using too many DFVs will not only increase cost, but also deliver noisy information, so we need to find an optimal number. In this experiment, we adopted the VGG-based T-DNN used in Section 4.2. We took into account two types of S-DNNs for this experiment: S-DNN with pooling and S-DNN with stride.
    
    The experimental results of the proposed method were shown in Table 5. In general, performance was improved regardless of the number of DFVs, but in the case of S-DNN with pooling, we could observe that as the number of DFVs becomes too large, the accuracy rises and drops again. This is because the distillation of too much amount of knowledge may cause transfer of even unnecessary information as mentioned in Section 3. However, S-DNN with stride shows a slight increase in performance. This is because the performance of the S-DNN is relatively low compared to that of the T-DNN, so receiving additional knowledge will significantly improve performance. Therefore, a reasonable number of DFVs should be used depending on the available cost, and the number of DFVs required can be determined according to the structure of the network.
    
    \begin{table}[t]
    \begin{center}
    \caption{Performance comparison according to the number of DFVs.}
    \label{table:headings}
    \begin{tabular}{c|c|cccccc}
    \hline\hline\noalign{\smallskip}
    \multirow{2}{*}{VGG} & \multirow{2}{*}{Model} & \multicolumn{6}{c}{The number of DFVs}\\
                                                 && - & 1 & 2 & 4 & 8 & 16\\
    \noalign{\smallskip}
    \hline\hline
    \noalign{\smallskip}
    \multirow{2}{*}{VGG} & S-DNN w/ pool   & 61.37 & 65.54 & \textbf{66.33} & 66.17 & 65.38 & 65.15 \\
                         & S-DNN w/ stride & 54.17 & 61.28 & 61.54 & 61.63 & 61.82 & 62.00\\

    \hline
    \end{tabular}
    \end{center}
    \end{table}

\section{Conclusion and Future Work}
    We propose a novel knowledge distillation method in this paper. The existing knowledge transfer technique 1) was limited to a limited network structure, 2) the quality of knowledge was low, and 3) as the learning progresses, the knowledge of the T-DNN vanished rapidly. We have proposed a method to transfer very rich information by defining novel knowledge using SVD and RBF, which are frequently used in traditional machine learning, without any structural limitations of the network. In addition, self-supervised learning associated with multi-task learning have been applied so that it was able to continue to receive T-DNN's knowledge during the learning process, which could also lead to additional performance enhancement. Experimental results showed that the proposed method has a significant improvement of about 4.96\% compared to the 3.17\% improvement in terms of accuracy performance based on VGG network \cite{yim2017gift}. In the future, we will develop a semi-supervised learning scheme by extending self-supervised learning concept through proposed knowledge transfer.
\linebreak

\noindent \textbf{Acknowledgements}: 
    This research was supported by National Research Foundation of Korea Grant funded by the Korean Government (2016R1A2B4007353).
    
\bibliographystyle{splncs}
\bibliography{egbib}
\end{document}